\documentclass[letterpaper]{article} % DO NOT CHANGE THIS
\usepackage{aaai2026}  % DO NOT CHANGE THIS
\usepackage{times}  % DO NOT CHANGE THIS
\usepackage{helvet}  % DO NOT CHANGE THIS
\usepackage{courier}  % DO NOT CHANGE THIS
\usepackage[hyphens]{url}  % DO NOT CHANGE THIS
\usepackage{graphicx} % DO NOT CHANGE THIS
\usepackage{natbib}  % DO NOT CHANGE THIS AND DO NOT ADD ANY OPTIONS TO IT
\usepackage{caption} % DO NOT CHANGE THIS AND DO NOT ADD ANY OPTIONS TO IT
\usepackage{algorithm}
\usepackage{algorithmic}
\usepackage[table]{xcolor}

\usepackage{newfloat}
\usepackage{listings}
\DeclareCaptionStyle{ruled}{labelfont=normalfont,labelsep=colon,strut=off} % DO NOT CHANGE THIS
\floatstyle{ruled}
\newfloat{listing}{tb}{lst}{}
\floatname{listing}{Listing}
\title{Ideal Registration? Segmentation is All You Need}

\author{
    Xiang Chen\textsuperscript{\rm 1},
    Fengting Zhang\textsuperscript{\rm 1},
    Qinghao Liu\textsuperscript{\rm 1},
    Min Liu\textsuperscript{\rm 1}\thanks{Corresponding authors:liu\_min@hnu.edu.cn},
    Kun Wu\textsuperscript{\rm 2},
    Yaonan Wang\textsuperscript{\rm 1},
    Hang Zhang\textsuperscript{\rm 3}
}

\affiliations{
    \textsuperscript{\rm 1}Hunan University,  
    \textsuperscript{\rm 2}Shanghai Normal University,  
    \textsuperscript{\rm 3}Cornell University\\
}

\usepackage{bibentry}
\begin{document}

\maketitle

\begin{abstract}
Deep learning has revolutionized image registration by its ability to handle diverse tasks while achieving significant speed advantages over conventional approaches. Current approaches, however, often employ globally uniform smoothness constraints that fail to accommodate the complex, regionally varying deformations characteristic of anatomical motion. 
To address this limitation, we propose \textbf{SegReg}, a ~\textbf{Seg}mentation-driven ~\textbf{Reg}istration framework that implements anatomically adaptive regularization by exploiting region-specific deformation patterns.
Our SegReg first decomposes input moving and fixed images into anatomically coherent subregions through segmentation. These localized domains are then processed by the same registration backbone to compute optimized partial deformation fields, which are subsequently integrated into a global deformation field. 
SegReg achieves near-perfect structural alignment (98.23\% Dice on critical anatomies) using ground-truth segmentation, and outperforms existing methods by 2-12\% across three clinical registration scenarios (cardiac, abdominal, and lung images) even with automatic segmentation. Our SegReg demonstrates a near-linear dependence of registration accuracy on segmentation quality, transforming the registration challenge into a segmentation problem. The source code will be released upon manuscript acceptance. %Building upon recent segmentation advances, SegReg establishes new performance standards without compromising computational efficiency. The source code will be released upon manuscript acceptance.
\end{abstract}


\section{Introduction}
Deformable image registration is a fundamental task for various medical imaging tasks such as surgical planning, surgical navigation, disease diagnosis and motion analysis \cite{chen2021deepSurvey,viergever2016survey}, which aims to establish dense, non-linear correspondences (deformation fields) between the moving and fixed images. %Given input moving and fixed images, the registration task is to find a deformation field/matrix which can align the moving image into the fixed image. %Therefore, different from other visual tasks like segmentation and classification, image registration naturally do not have the ground-truth deformation fields, thereby limiting the application of those supervised learning-based approaches. 

\begin{figure}[t]
    \centering
    \includegraphics[width=1.0\linewidth]{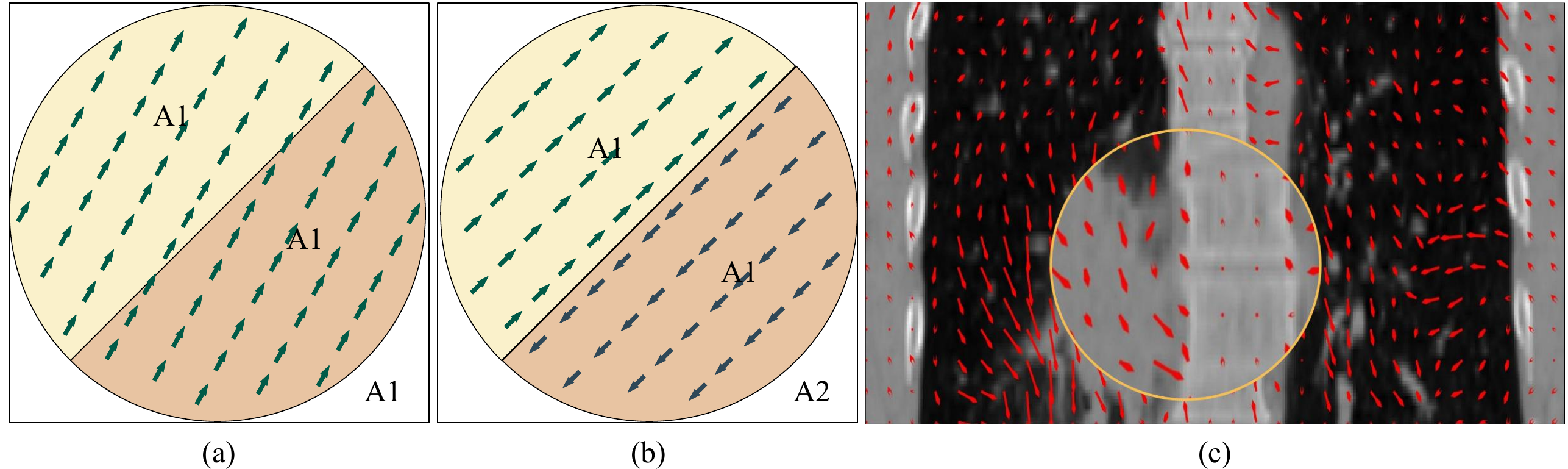}
    \caption{
    Visualization of different motion assumptions. Most learning-based registration methods assume totally smooth deformation fields (Assumption A1). Instead, SegReg assumes deformation fields to be smooth within each sub-region while discontinuous on the boundaries (Assumption A2), accommodating more complex scenarios like (c).
    }
    \label{fig:concept_fig}
    \vspace{-2ex}
\end{figure}

Traditional methods \cite{beg2005computing,ashburner2007fast,vercauteren2009diffeomorphic,avants2011reproducible,marstal2016simpleelastix} typically employ a multi-step optimization strategy aimed at minimizing an error function comprising similarity and smoothness regularization. The similarity metric quantifies the discrepancy between the warped moving image and the fixed image, whereas smoothness regularization enforces a smooth deformation field, thereby mitigating irregular motions. 
%However, the imposition of smoothness regularization can occasionally compromise registration accuracy by oversimplifying the deformation model.
Although these methods are reliable for accurate registration, their time-intensive process is a notable limitation in scenarios demanding rapid registration~\cite{chen2021deepSurvey}. 

Deep learning registration methods have seen broad application, offering efficient, near-real-time registration post-training\cite{balakrishnan2019voxelmorph,chen2022transmorph}. Unsupervised registration networks\cite{balakrishnan2019voxelmorph}, which utilize solely the moving and fixed images as inputs, have matched the performance of conventional approaches. 
They typically employ similarity metrics such as Mean Squared Error (MSE) and Normalized Cross Correlation (NCC), coupled with smooth regularization techniques like L2 diffusion, for network training. To enhance registration performance, weakly supervised registration networks incorporate supplementary segmentation masks or landmarks as additional loss functions like Dice loss to refine the guidance \ cite {chen2021deepDiscontinuity}. 
Despite the escalating complexity of registration models in recent years, notably with the incorporation of transformers \cite{chen2022transmorph}, large convolutional kernels \cite{jia2022u} and pyramid-based registration \cite{zhang2024memwarp, wang_RDP}, the enhancements in registration accuracy have reached a plateau. In addition, \cite{jena2024deep} points out that, architectural designs in learning-based methods are unlikely to affect the registration performance, without the guidance of weak supervision information. Therefore, additional label information is imperative to unlock the full potential of learning-based approaches and realize substantial advancements in the precision of registration techniques.

To our analysis, a primary constraint on current registration accuracy is the smooth regularization inherent in most deep learning-based registration networks~\cite{balakrishnan2019voxelmorph,chen2022transmorph}, which assumes \textbf{the deformation fields are globally smooth (Assumption A1)}. While valid for simple tasks (e.g., single-object alignment or local minor physiological shifts), this paradigm artificially constrains performance in anatomically heterogeneous scenarios due to overly constrained regularization. As shown in Figure \ref{fig:concept_fig}, scenarios (a) and local regions in (b) and (c) align with Assumption A1, whereas the complete images of (b) and (c) do not. In medical imaging, the diverse properties and motions of organs and tissues are poorly captured by global smooth regularization, which hinders the accurate alignment of disparate organs or tissues. To better approximate realistic motion and enhance registration performance, discontinuity-preserving registration techniques~\cite{ng2020unsupervised,chen2021deepDiscontinuity} are necessary, which assumes \textbf{the deformation fields to be locally smooth and globally discontinued (Assumption A2)}, achieved via structure guidance. 

Contrary to registration, segmentation ground truth (GT) is more accessible, with numerous public segmentation models available \cite{wasserthal2023totalsegmentator,isensee2021nnu}. Despite scanner and scanning condition variations, images of identical organs and modalities share a high degree of similarity. Consequently, with limited local data, high segmentation accuracy can be attained by finetuning these public segmentation models. The emergence of nnUNet \cite{isensee2021nnu} and Segment Anything Model (SAM \cite{kirillov2023segment,chen2024ma,huang2024segment}) has streamlined effective segmentation across diverse tasks, even with limited training data. Building on these strategies, enhancing registration performance through precise segmentation seems a logical next step.

In this paper, we embrace the principles of discontinuity-preserving registration (\textbf{Assumption A2}), assuming: (1) the motion within organs/regions is smooth and consistent in the absence of external factors, and (2) motion between different organs/regions could be either continuous or discontinuous, where the discontinuity occurs at the boundaries between different organs/regions. This assumption, while simple, encompasses a wider range of motion patterns than the traditional uniform smoothness assumption. Leveraging this, optimal registration within each sub-region is achievable through a simple end-to-end network under \textbf{Assumption A1}. %With accurate segmentation more feasible, we propose to exceed the accuracy limits of smooth registration methods through region-wise registration. 
We introduce SegReg, a novel framework that transforms registration into a segmentation task, comprising region-wise mapping and deformation field composition. Region-wise mapping aligns corresponding subregions in moving and fixed images, while composition integrates subregional deformation fields into a complete image field. Although the composition scheme naturally captures discontinuities, it may sometimes overemphasize boundary discontinuities. To ensure smooth transitions between subregions, we further employ an Euclidean distance transform (EDT) loss in addition to the standard Dice loss.
The contributions are summarized as follows:
\begin{itemize}
    \item We propose a novel registration framework, SegReg, which transfers the complex registration problem into a couple of simple region-wise registrations, via segmentation, naturally preserving discontinuities and breaking the upper limit of current registration approaches.
    \item We introduce EDT loss, which employs different weights for segmentation masks within each region, enabling a smoother transfer across region boundaries while keeping discontinuity.
    \item Results on three different registration tasks demonstrate that the proposed SegReg framework significantly outperforms other state-of-the-art (SOTA) methods. %SegReg also archives the Rank 2 place in the ThoraxCBCT \cite{hugo2017longitudinal} leaderboard from Learn2Reg2023.
\end{itemize}

\section{Related Work}
%According to the available data in the network training, learning-based registration networks can be divided into three classes: supervised registration, unsupervised registration and weakly-supervised registration.

\subsection{Supervised Registration Methods}

Unlike other visual tasks, image registration lacks definitive GT deformations. Given the inaccessibility of true deformation fields, early deep registration methods relied on synthetic deformations or those predicted by traditional approaches. Supervised methods train end-to-end networks to predict deformation fields by minimizing the discrepancy between predictions and reference deformations. For instance, similarity-steered regression \cite{cao2017deformable} uses SyN \cite{avants2008symmetric} and Diffeomorphic Demons \cite{vercauteren2009diffeomorphic} to generate GT deformation fields. Similarly, BIRNet \cite{fan2019birnet} uses existing methods' deformations and trains the network alongside intensity dissimilarity measure. SVF-Net \cite{rohe2017svf} constructs target fields by matching shapes \cite{durrleman2014morphometry} from segmentations. %Quicksilver \cite{yang2017quicksilver} initializes LDDMM with predicted momenta for acceleration. 
In addition, synthetic transformations, including affine and elastic deformations \cite{eppenhof2018pulmonary,young2022superwarp}, are integrated and composed across scales to create GT fields.

\subsection{Unsupervised Registration Methods}
%In this paper, the term "self-supervised registration" collectively includes unsupervised learning, semi-supervised learning, and weakly supervised learning, defined by the absence of direct supervision of target warp.
For unsupervised registration, the available data for network training are only the input images. To train the network, an end-to-end network takes moving and fixed images as input and predicts a corresponding deformation field. A spatial transformer block warps moving images to approach fixed images. The training loss comprises two parts: the similarity between warped moving image and fixed image, and the smooth regularisation of deformation fields.
VoxelMorph \cite{balakrishnan2019voxelmorph, dalca2019unsupervised} led the way in unsupervised registration, showing a learning-based method can match traditional iterative methods' results in under a second for volumetric brain scans. Subsequently, numerous networks have been developed to enhance the accuracy of registration, smoothness of deformation fields or computational efficiency. In architecture design, studies have investigated parallel \cite{kang2022dual,jia2022u} and cascaded structures \cite{zhao2019recursive, mok2020large}, attention-based transformers \cite{chen2022transmorph,shi2022xmorpher} along with multi-layer perceptrons \cite{meng2024correlation} for enhanced learning. For enhancing computational efficiency, DeepFlash \cite{wang2020deepflash} and Fourier-Net \cite{jia2023fourier,jia2023fourierplus} approximate the full deformation field with a band-limited version, while approaches such as LessNet \cite{jia2024decoder} with a decoder-only configuration and the Slicer Network \cite{zhang2024slicer} with an encoder-only setup reduce redundancy and boost efficiency.

% Further research has explored various network designs, including parallel \cite{kang2022dual,jia2022u} and cascaded architectures \cite{zhao2019recursive, mok2020large, jia2021learning}, as well as the incorporation of attention-based transformers \cite{chen2022transmorph} to refine the representation learning. 

\subsection{Weakly-supervised Registration}
%Research in weakly-supervised registration proposes to utilise additional supervised information like segmentation masks or landmarks in the network training, comprising two types: (1) only use additional supervised information in the training process. (2) utilize the supervised information in both training and testing. The first type generally utilizes the segmentation masks or landmarks as an additional similarity loss such as Dice loss. In this scenario, the unsupervised registration networks can be directly trained in a weakly-supervised manner, resulting in higher registration accuracy. Other key advancements include new loss functions like imposing local-oriented consistency \cite{mok2020fast} for diffeomorphism \cite{dalca2019unsupervised}, auxiliary segmentation for anatomical alignment \cite{balakrishnan2019voxelmorph, mok2021large}, and regularizers for curvature \cite{hering2021cnn} and keypoints \cite{hering2021cnn,zhang2024slicer}. For the second type, a typical architecture is DDIR~\cite{chen2021deepDiscontinuity}, which built a four-channel encoder-decoder for cardiac image registration, where the segmentation masks are used to split the original image into four different pairs. Similarly, Chen et al.~\cite{chen2024textscf} utilized the segmentation of fixed images and large language model-enabled text-encoder to build a text spatially covariant filter, and demonstrated that even low-accuracy automatic segmentation can augment the resultant registration performance.

Weakly-supervised registration research incorporates additional supervision, such as segmentation masks or landmarks, in network training, falling into two categories: (1) using this information solely during training, and (2) employing it in both training and testing phases. The first approach often integrates these cues as an auxiliary similarity loss, like Dice loss \cite{zhang2024memwarp}, enhancing unsupervised networks via weakly-supervised training and boosting registration precision. Notable developments include novel loss functions that impose local-oriented consistency \cite{mok2020fast} for diffeomorphism \cite{dalca2019unsupervised}, auxiliary segmentation for anatomical alignment \cite{balakrishnan2019voxelmorph, mok2021large}, and regularizers for curvature \cite{hering2021cnn} and keypoints \cite{hering2021cnn,zhang2024slicer}. 
The second approach, exemplified by DDIR~\cite{chen2021deepDiscontinuity}, constructs a multi-channel encoder-decoder for cardiac registration, leveraging segmentation masks to segment the image into distinct pairs, left ventricle blood pool (LVBP), left ventricle myocardium (LVM), right ventricle (RV) and background, respectively, and conducting region-wise registration. Similarly, Chen et al.~\cite{chen2024textscf} utilized fixed image segmentation and a text-encoder from large language models to create a spatially covariant filter, showing that even low-accuracy automatic segmentation can significantly enhance registration.

%Owing to the scarcity of labeled data, previous studies have generally shied away from incorporating segmentation within the registration network. Even when employing weakly-supervised training for registration networks, the tendency has been to leverage segmentation solely for the Dice loss, thus circumventing the necessity for segmentation during inference. However, absent the guidance of labelled information, achieving discontinuity-preserving registration is challenging, which in turn limits the accuracy of registration. With the advancement of publicly available segmentation models and large-scale visual segmentation models, accurate segmentation is now more accessible. Our SegReg framework bridges segmentation and registration, offering a potential solution to surpass the current upper bounds of registration accuracy.

Due to limited labelled data, previous research has often avoided integrating segmentation into registration networks. Despite using weakly-supervised training, segmentation was typically used only for Dice loss, bypassing its need during inference. These approaches, lacking labelled guidance in the inference, hindered the achievement of discontinuity-preserving registration and restricted registration accuracy. With the progress in public segmentation models and foundation segmentation models, accurate segmentation has become more feasible. Our SegReg bridges segmentation and registration, surpassing current registration accuracy limits.

%However, with the progress in public segmentation models and large-scale visual segmentation, accurate segmentation has become more feasible. Our SegReg framework now bridges segmentation and registration, potentially surpassing current registration accuracy limits.

\section{Methodology}
%In this section, we first introduce the framework architecture of SegReg, and then discuss the corresponding loss function and training/inference strategies.

\begin{figure*}[htb]
\centering
\includegraphics[width=\linewidth]{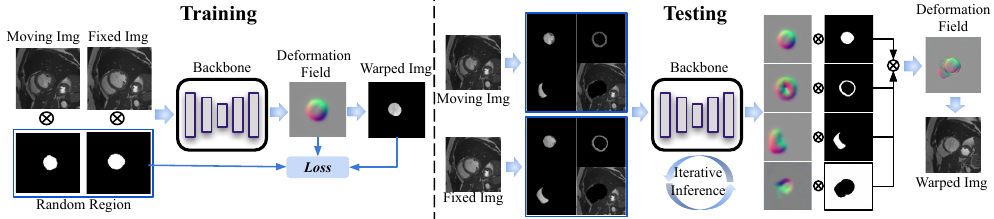}
\caption{Schema of SegReg. The input moving and fixed images are first split into different pairs of interest using the segmentation masks. In the training process, a random pair is fed into the registration backbone at each iteration, and update the parameters of the registration backbone via back-propagation. In the test phase, all pairs of interest are fed into the registration backbone and predict the corresponding sub-deformation fields, and then combined to obtain the final deformation fields.
}
\label{fig:network}
\vspace{-2ex}
\end{figure*}

\subsection{Framework of SegReg}
%The overall framework of SegReg can be found in Figure~\ref{fig:network}. The main idea of SegReg is to warp the moving image to the fixed image by aligning each local region. To do this task, segmentation masks are required, either manual segmentation or automatic segmentation. The automatic segmentation masks can be obtained using the same training data for the network training, or finetuning a publicly available segmentation model (like SAM). The deformation of each local pair is obtained by a registration backbone, where the predicted deformation fields are sub-deformation fields, corresponding to the sub-regions. During the training, we randomly select a pair of regions from the original moving and fixed images, and train the registration backbone as a regular weakly-supervised registration network. In the inference phase, all pairs of regions are fed into the registration backbone to achieve the sub-deformation fields, and compose them into the complete deformation fields. The warped moving image is then obtained by warping the original moving image with the complete deformation fields.

The overall framework of SegReg is depicted in Figure~\ref{fig:network}. The core concept of SegReg is to match the moving image and fixed image by aligning each local region. For this process, segmentation masks are essential, which can be sourced from either manual annotations or automated segmentation techniques. Automated segmentation masks can be derived from the same dataset used for network training or by finetuning a pre-existing segmentation model, such as SAM. Each local pair's deformation is calculated by a registration backbone, yielding sub-deformation fields that correspond to individual sub-regions. During training, we randomly select pairs of regions from the original moving and fixed images, training the registration backbone in a manner akin to a standard weakly-supervised registration network. In the inference, all pairs of regions are processed by the registration backbone to predict sub-deformation fields, which are then composed into overall deformation fields. The warped image is subsequently generated by applying overall deformation fields to the original moving images.

Denote the moving image as $I_m$ and fixed image as $I_f$, the process of image registration can be formulated as,
\begin{equation}
    \mathbf{\Phi}(x) = x + \mathbf{u}(x),
\label{eq:deformation}
\end{equation}
\begin{equation}
    I_f = \mathbf{\Phi}(x) \circ I_m,
\label{eq:def}
\end{equation}
where the $\mathbf{u}(x)$ is the displacement field, denoting the movement (x,y for 2D image registration, and x,y,z for 3D image registration) of each voxel in the images. The $\mathbf{\Phi}()$ is the warping function from the moving image to the fixed image.
For n-label images, denote the moving and fixed segmentation masks as $S_m$ and $S_f$, then $S_m = \{{S_m}^0,..., {S_m}^{n-1}\}$, $S_f = \{{S_f}^0,..., {S_f}^{n-1}\}$. Within each sub-region, the deformation field is smooth. In the training, we can rewrite the formula \ref{eq:def} for i-th region as,
\begin{equation}
    I_f \times {S_f}^i = \mathbf{\Phi_i}(x) \circ (I_m \times {S_m}^i), i \in \{0,...,n-1\}.
\label{eq:sub_def}
\end{equation}

%Given input moving image $I_m$ and fixed image $I_f$, the task of image registration is to estimate a displacement field $u(x)$ which can warp the moving image as close as possible to the fixed image. Following~\cite{ashburner2007fast}, the wrapping function of small deformation can be formulated as,

\subsection{Registration Backbone}
\begin{figure}[ht]
\centering
\includegraphics[width=\linewidth]{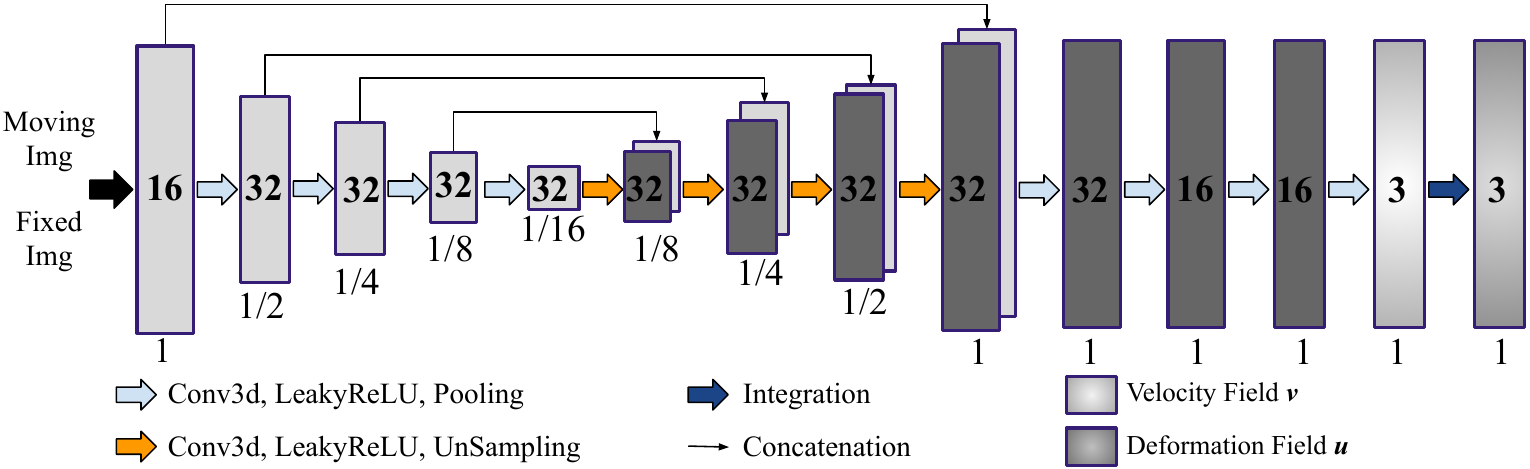}
\caption{Architecture of registration backbone in SegReg. For simplicity, we utilize a general UNet as the backbone.
}
\label{fig:registration_backbone}
%\vspace{-2ex}
\end{figure}

To demonstrate the efficiency of our proposed SegReg framework, we choose a simple UNet as the registration backbone, following~\cite{balakrishnan2019voxelmorph, dalca2019unsupervised, chen2021deepDiscontinuity}. As shown in Figure \ref{fig:registration_backbone}, the registration backbone comprises an encoder and a decoder. The encoder consists of four layers that successively downsample the input images by half with each layer. The decoder, mirroring the encoder, comprises four layers that upsample the feature maps to double their original size. Skip-connection is applied on the corresponding layers in the encoder and decoder. The moving and fixed regions are concatenated at the beginning, and then fed into the encoder-decoder to predict the velocity field $\textbf{v}$. The deformation fields $\textbf{u}$ are obtained by integration from the velocity fields $\textbf{v}$, following \cite{dalca2019unsupervised}.
Note that, under our SegReg framework, even a simple UNet is enough to obtain superior registration accuracy, while it can also be replaced by more efficient or powerful architectures, to further improve registration performance.

%It is noteworthy that while the registration backbone can be substituted with more sophisticated architectures, the registration accuracy, under the SegReg framework, remains comparable, nearing an impressive 100\% accuracy when guided by ground-truth segmentation masks. This highlights the robustness of SegReg in achieving superior registration results with various backbone configurations.

\subsection{Deformation Field Composition}
%As the motion within a sub-region is smooth and continuous, a simple registration backbone conforming to the global smooth assumption can predict superior deformation in each sub-region.
%However, it also brings a problem: the resultant deformation fields only correspond to local regions, unable to tackle the deformation of the complete image. Therefore, in the test phase, a composition scheme of the sub-deformation fields for each region in the original input is required. 
Given that motion within a sub-region is smooth, a straightforward registration backbone that adheres to Assumption A1 can effectively predict deformation within each sub-region. 
Nevertheless, this strategy introduces a challenge: the derived deformation fields are localized and do not account for the global deformation. Consequently, in the testing, a composition scheme is required to integrate the sub-deformation fields from each region into complete deformation fields. %This compositional approach is necessary to address the deformation across the entire image, ensuring that the registration process is not confined to isolated regions but rather considers the image as a comprehensive entity.
Denote the deformation field corresponding to segmentation mask i as: $\mathbf{\Phi}_i$, the deformation field composition process is formulated as $\mathbf{\Phi}(x) =  \sum_{i}^{n}(\mathbf{\Phi_i}(x)) \times S_i$. Once achieving the $\mathbf{\Phi}$, the final warped images are achieved by warping moving images with $\mathbf{\Phi}$.

% \begin{equation}
%     \mathbf{\Phi}(x) =  \sum_{i}^{n}(\mathbf{\Phi_i}(x)) \times S_i.
% \label{eq:composiiton}
% \end{equation}
% Once achieving the $\mathbf{\Phi}$, the final warped moving image is achieved by warping the original image with $\mathbf{\Phi}$.

%The strategy of splitting the original images and composing the resultant sub-deformation fields brings the following advantages: 
The split-and-merge strategy offers two key advantages: (1) By aligning corresponding local regions in isolation, we can capture the optimal smooth registration within each sub-region, leveraging the inherent continuity of motion within localized areas. (2) The amalgamation of these smooth sub-deformation fields into complete deformation fields ensures that the final deformation is locally smooth while global discontinuities. This naturally aligns with the principles of discontinuity-preserving registration, effectively accommodating varying motion patterns across different regions.
%Splitting the original images into different pairs and registering one pair each iteration can save the model parameters, enabling those complex anatomical structures with a lot of segmentation masks. (3) Composition of smooth sub-deformation fields ensures that the final deformation fields are locally smooth while globally discontinuity, consistent with realistic motion (e.g. on sliding scenarios).

%to do
\subsection{Loss Function}
To optimize our SegReg, we employ a hybrid loss function strategy that integrates two components: a similarity loss and a regularization loss. The similarity loss is designed to assess and quantify the alignment between the warped moving image and the fixed image. On the other hand, the regularization loss is instrumental in enforcing the smoothness of the predicted deformation fields, which is crucial for ensuring the physiological plausibility of the transformations.

\begin{figure}[ht]
\centering
\includegraphics[width=0.8\linewidth]{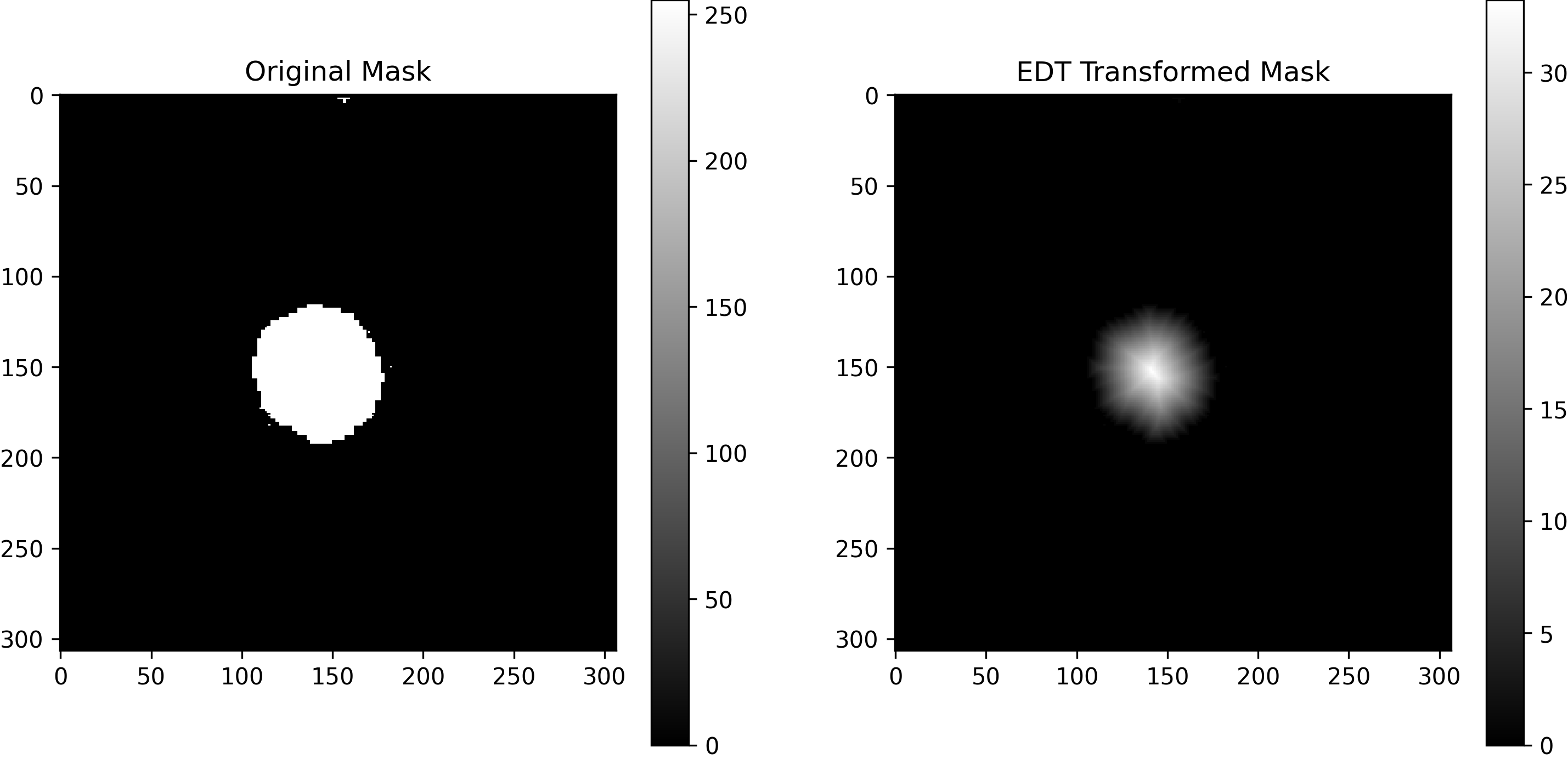}
\caption{Segmentation mask and corresponding EDT mask. 
}
\label{fig:edt_loss}
%\vspace{-2ex}
\end{figure}

The similarity loss further comprises two parts, a voxel-wise loss ${L}_v$ and a mask-wise loss ${L}_m$. The voxel-wise loss can be NCC or MSE loss. Regarding the mask-wise loss, a popular choice could be the Dice loss, which helps to align the images according to regions and improve registration accuracy. However, in the SegReg framework, it may cause severe gradient mutation across the boundaries, leading to significant discontinuity on the boundary. To alleviate the phenomenon, we further propose an EDT loss which transfers the original binary segmentation masks (in training, the input registration pair only contains the foreground and background) into EDT masks, as shown in Figure \ref{fig:edt_loss}. Different from binary masks, the pixels/voxels in the foreground turn to their distance to the boundaries (normalized to 0-1), thereby implying different weights to pixels/voxels and surpassing those pixels/voxels on the boundary. Then the EDT loss is computed as the MSE loss between the warped moving EDT masks and fixed EDT masks, formulated as,
\begin{equation}
\label{eqn:EDT}
L_{edt} = \sqrt{({EDT}_f - \mathbf{\Phi_i}(x) \circ {EDT}_m)^2}.
\end{equation}

According to our assumption, the deformation fields are smooth and continuous within each sub-region. In the training phase, we only register one pair of regions each iteration. Following previous research~\cite{dalca2019unsupervised,balakrishnan2019voxelmorph}, we directly choose $L_2$-regularisation on each deformation field as the smooth regularisation $R$. Therefore, the total loss $L_{total}$ is formulated as,
\begin{equation}
\label{eqn:loss_total}
L_{total} =  \gamma_0 \times {L}_v + \gamma_1 \times L_{m} + \gamma_2 \times R,
\end{equation}
where the $\gamma_0$, $\gamma_1$, $\gamma_2$ are hyperparameters to balance the similarity loss and regularisation, set empirically.

%\begin{equation}
%\label{eqn:regularisation}
%R =  ||\nabla \textbf{u}||^2.
%\end{equation} 

\subsection{Training and Inference}
%The SegReg framework can be applied either as a learning-based registration network, or as an iterative registration method like traditional approaches. For learning-based registration, the registration pair (moving and fixed images) are first split into different regions of interest, using the corresponding segmentation masks. In the network training, SegReg randomly input a pair of local regions into the registration backbone each iteration. In the testing phase, all pairs of regions in segmentation masks would be fed into the registration backbone and achieve the corresponding sub-deformation fields, followed by a composition operation to obtain the final deformation fields. The final warped image is obtained by deforming the original moving image via the final deformation fields.
The SegReg framework's adaptability enables its utilization in dual capacities: as a learning-based registration network or as an iterative registration technique, mirroring conventional approaches. For learning-based registration, the registration pair is initially segmented into various regions of interest, guided by the segmentation masks. During training, SegReg randomly introduces a random pair of local regions into the registration backbone at each iteration. In the inference, every pair of regions identified by the segmentation masks is fed to the registration backbone to generate corresponding sub-deformation fields, which are then combined by a composition operation to yield the complete deformation fields. The warped image is obtained by applying the complete deformation field to the original moving images, thus facilitating precise alignment with fixed images.

For iterative registration, SegReg is similar to traditional iterative registration methods, which register the moving and fixed images via multiple iterations and do not require a training stage prior to the inference. Within each iteration, the deformation fields are refined based on a hybrid loss function that encompasses both similarity metrics and smooth regularization. Instead of using global smooth regularization, we compute the summary of $L_2$-regularization within each sub-region (split by the corresponding fixed segmentation) as the total smooth regularisation $R$.

\section{Experiments \& Results}

%In this section, we assess the performance of our proposed SegReg framework against leading image registration techniques using three datasets that encompass diverse imaging modalities, input constraints, and anatomical structures. We first provide the detailed settings of the experiments, then describe the corresponding experiment results and analysis.

%Subsequent subsections elaborate on the datasets, implementation specifics, comparative baseline methods, and the evaluation criteria employed.
%We then present the qualitative and quantitative outcomes, showcasing registration efficacy across three distinct tasks. This also includes an analysis comparing ground-truth segmentation with automatic segmentation results, an examination of discontinuity preservation, and an assessment of the impact of segmentation variability on registration accuracy.

\subsection{Datasets and Implementation Details}

We demonstrate our SegReg framework on both computed tomography (CT) and magnetic resonance (MR) image modalities, with inter-subject and intra-subject settings.
Specifically, we use the ACDC dataset \cite{bernard2018deep} for intra-subject cardiac cine-MR image registration, the Abdomen CT dataset for inter-subject registration \cite{xu2016evaluation}, and ThroaxCBCT dataset \cite{hugo2017longitudinal} for intra-subject multi-modality thorax image registration. More details about the datasets can be found in the Appendix.

In our experiments, all models were developed using PyTorch library in Python on an A6000 GPU machine. 
We utilize Adam optimizer for network training, with an initial learning rate of $1e-4$ and a polynomial learning rate scheduler with a decay rate of 0.9. MSE loss, NCC loss, and MSE loss on MIND features \cite{heinrich2012mind} are used as the voxel-wise loss $L_v$ in cardiac images, abdomen images, and thorax images, respectively. We build three versions of SegReg, SegReg-Dice, SegReg-EDT and SegReg-EDT\&Dice, using Dice loss, EDT loss and the hybrid loss of them ($L_m = (L_{Dice} + 100\times L_{EDT})/2$) as the mask-wise loss $L_m$ in network training. In cardiac image registration, we set $\gamma_0:\gamma_1:\gamma_2 = 1:0.1:0.01$ for SegReg-Dice and SegReg-EDT\&Dice, and $\gamma_0:\gamma_1:\gamma_2 = 1:10:0.01$ for SegReg-EDT.
In abdominal image registration, we choose $\gamma_0:\gamma_1:\gamma_2 = 1:1:0.1$ for SegReg-Dice and SegReg-EDT\&Dice, $\gamma_0:\gamma_1:\gamma_2 = 1:100:0.1$ for SegReg-EDT.
%The number of time steps in integration is set to 7. For a fair comparison, all models were trained under identical conditions or, where necessary, using the optimal settings outlined in their original publications.
%Dataset-specific details, including the dissimilarity function $s(\cdot)$, inclusion of Dice loss, and other training parameters, are provided in the appendix.

%Swin-UNETR~\cite{tang2022self,hatamizadeh2022swin} (https://huggingface.co/darragh/swinunetr-btcv-base) and nnUNet~\cite{isensee2021nnu}(https://zenodo.org/records/3734294)
%to do
%\subsubsection{Baseline Methods}
We compare our SegReg framework with SOTA learning-based registration models. All the approaches are trained in a weakly-supervised manner, trained with Dice loss. For a fair comparison, we employ two publicly available segmentation models: Swin-UNETR~\cite{tang2022self,hatamizadeh2022swin} and nnUNet~\cite{isensee2021nnu}, to segment abdominal CT images and cardiac MR images, respectively, and then feed the resultant predictions into our SegReg for inference. We utilize their pre-trained models directly from online sources, \textbf{without any local fine-tuning}. The segmentation accuracy on the abdominal and ACDC test sets are 75.83\% and 91.19\%, respectively. For ThoraxCBCT, we obtain results directly from the public leaderboard. % or respective publications. In ACDC and Abdomen CT datasets, we follow publicly available codes for each model to achieve optimal performance.
%We compare our SegReg framework with several SOTA learning-based baseline models, including VoxelMorph \cite{balakrishnan2019voxelmorph}, TransMorph \cite{chen2022transmorph}, DDIR \cite{chen2021deepDiscontinuity}, LKU-Net \cite{jia2022u}, Fourier-Net \cite{jia2023fourier}, RDP~\cite{wang_RDP}, Xmorpher \cite{shi2022xmorpher}, and MemWarp \cite{zhang2024memwarp}. For the Abdomen CT dataset, we additionally include textSCF \cite{chen2024textscf}, LapIRN \cite{mok2021large}, and discrete optimization-based methods ConvexAdam \cite{siebert2021fast} and SAMConvex \cite{li2023samconvex} in the comparison, as these are highly effective for handling large deformations. Note that, all the approaches are trained in a weakly-supervised manner, using Dice loss to guide the network training. For a fair comparison, we employ two publicly available segmentation models: \url{https://huggingface.co/darragh/swinunetr-btcv-base}{Swin-UNETR}~\cite{tang2022self,hatamizadeh2022swin} and \url{https://zenodo.org/records/3734294}{nnUNet}~\cite{isensee2021nnu}, to segment abdominal CT images and cardiac MR images, respectively, and then feed the resultant predictions into our SegReg framework for inference. We utilize their pre-trained models directly from online sources, \textbf{without any local fine-tuning}. The segmentation accuracy on the abdominal and ACDC test sets are 75.83\% and 91.19\%, respectively. For results on ThoraxCBCT, we obtain evaluation scores from the public leaderboard or respective publications. In ACDC and Abdomen CT datasets, we follow publicly available codes for each model to achieve optimal performance.

%\subsubsection{Evaluation Metrics}
Following previous research \cite{balakrishnan2019voxelmorph} and challenge protocols \cite{hering2022learn2reg}, we evaluate anatomical alignment using the Dice Similarity Coefficient (Dice) and the 95\% Hausdorff Distance (HD95). %For results on ThoraxCBCT, target registration error (TRE) is also evaluated.
To assess field smoothness, we measure the standard deviation and number of foldings of the logarithm of the Jacobian determinant (SDlogJ and J\textless 0). % and the number of negative Jacobian determinant voxels (J\textless 0). 
Computational complexity is assessed in cardiac image registration, using the multiply-add operations (MAs) and total parameter size (PS).

\subsection{Results and Analysis}
\subsubsection{Intra-subject Registration on ACDC.} 
We assess the intra-subject registration performance of SegReg on cardiac MR images from the ACDC dataset, as shown in Table \ref{tab:ACDC} and Figure \ref{fig:acdc_vis} (Boxplot can be found in the Appendix Figure \ref{fig:acdc_boxplot}). 
Leveraging their capacity to manage large deformations, pyramid-based methods MemWarp \cite{zhang2024memwarp} and RDP \cite{wang_RDP} demonstrate superior performance over conventional single-resolution deformation field prediction approaches. While achieving significantly enhanced registration accuracy (p$<$0.05), these methods incur prohibitively high computational overhead. 
As a discontinuity-preserving registration method also incorporating segmentation in the network, DDIR \cite{chen2021deepDiscontinuity} outperforms MemWarp \cite{zhang2024memwarp} and RDP \cite{wang_RDP}. Under SegReg framework, even a simple UNet registration backbone can outperform pyramid-based registration approaches. Our SegReg achieves superior registration accuracy compared to both MemWarp \cite{zhang2024memwarp}, RDP \cite{wang_RDP} and DDIR \cite{chen2021deepDiscontinuity}, while maintaining network complexity comparable to VoxelMorph \cite{balakrishnan2018unsupervised}, demonstrating the superiority.

\begin{table}[!t]
\begin{center}
\resizebox{1.\linewidth}{!}{
\begin{tabular}{ lcccccc }
\hline
\rowcolor{lightgray}
Model & Dice (\%) $\uparrow$ & HD95 (mm) $\downarrow$ &  SDlogJ $\downarrow$ &J(\textless 0) $\downarrow$ & MAs (G) $\downarrow$ & PS (MB) $\downarrow$ \\ 
\hline
Initial & 58.14 & 11.95 & - & - & - & -\\
\hline
VoxelMorph \cite{balakrishnan2018unsupervised} & 81.95& 7.18& \textbf{0.07}& 0.00 & 19.57& 0.33\\
TransMorph \cite{chen2022transmorph} & 82.75& 6.98& 0.084& 249.62&54.34& 46.69\\
LKU-Net \cite{jia2022u} & 82.37& 6.66& 0.08& 0.00 & 160.58& 33.35 \\
Fourier-Net \cite{jia2023fourier} & 83.98& 6.41& 0.07& 1.12 & 86.11& 17.43 \\
%Xmorpher \cite{shi2022xmorpher} & 78.59& 7.66& 0.07& 52.04&15.09\\
\hline
MemWarp \cite{zhang2024memwarp} & 87.81& 5.36& 0.08& 50.96 & 1267.28& 47.78 \\
RDP \cite{wang_RDP} & 85.99& 6.55& 0.086 & 0.00 & 154.14& 8.92 \\
DDIR \cite{chen2021deepDiscontinuity} & 88.03& 9.91& 0.121& 45.68 & 78.19& 1.31 \\
\hline
\textbf{SegReg-Dice(Ours)}& \textbf{89.98} & \textbf{5.08} & 0.36 & 41.52 & 19.57& 0.33\\
\textbf{SegReg-EDT(Ours)}& 87.79& 6.03 & 0.09 & 0.36 & 19.57& 0.33\\
\textbf{SegReg-EDT\&Dice(Ours)}& 89.24& 5.45 & 0.15 & 6.68 & 19.57& 0.33\\
\hline
\end{tabular}
}
\caption{
Quantitative comparison on the cardiac ACDC dataset. 
Best-performing metrics are highlighted in bold. 
Symbols indicate direction: $\uparrow$ for higher is better, $\downarrow$ for lower is better. 
``Initial" refers to results before registration.
%``MAs (G)" stands for multi-adds (G) and ``PS (MB)" is parameter size (MB).
}\label{tab:ACDC}
\end{center}
\vspace{-2ex}
\end{table}

\begin{figure*}[t]
    \centering
    \includegraphics[width=1.0\linewidth]{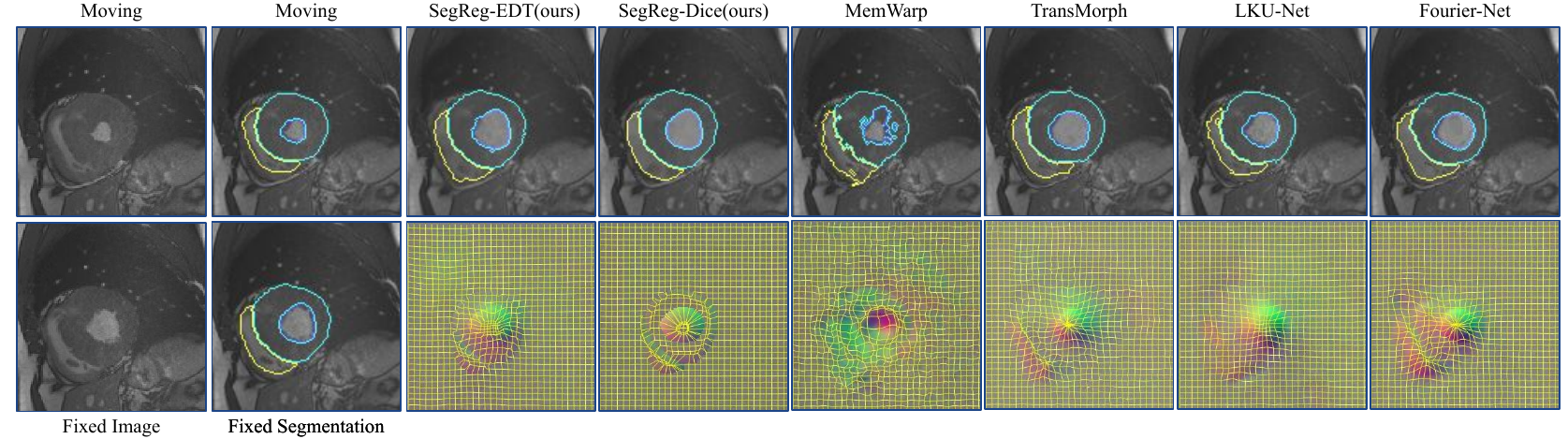}
    \caption{
    Qualitative comparison on ACDC. 
    SegReg-Dice achieves the best alignment while the discontinuity is severe on the boundary. In contrast, the SegReg-EDT can achieve a good balance of registration accuracy and smoothness.
    }
    \label{fig:acdc_vis}
    \vspace{-2ex}
\end{figure*}

\begin{table}[!t]
\begin{center}
\resizebox{\columnwidth}{!}{
\begin{tabular}{ lccc }
\hline
\rowcolor{lightgray}
Model &  Dice (\%) $\uparrow$ & HD95 (mm) $\downarrow$ & SDlogJ $\downarrow$ \\ 
\hline
Initial & 30.86 & 29.77 & - \\
% ConvexAdam~\cite{siebert2021fast} & 84.64 & 1.50 & \textbf{0.07} \\
\hline
VoxelMorph \cite{balakrishnan2018unsupervised} & 47.05 & 23.08 & 0.13\\
TransMorph \cite{chen2022transmorph} & 47.94 & 21.53 & 0.13 \\
LKU-Net \cite{jia2022u} & 52.78 & 20.56 & 0.98 \\
Xmorpher \cite{shi2022xmorpher} & 40.67& 23.77&0.14\\
LapIRN \cite{mok2020large} & 54.55 & 20.52 & 1.73 \\
%LessNet \cite{jia2024decoder} & 42.03 & 27.03 & \bf{0.07} \\
%FourierNet \cite{jia2023fourier} & 48.49 & 20.69 & 0.13 \\
Fourier-Net \cite{jia2023fourier} & 42.80 & 22.95 & 0.13 \\
ConvexAdam \cite{siebert2021fast} & 51.10 & 23.14 & \textbf{0.11}\\ 
SAMConvex \cite{li2023samconvex} & 53.65 & \textbf{18.66} & 0.12 \\
\hline
RDP \cite{wang_RDP} & 58.77 & 20.07 & 0.22 \\
MemWarp \cite{zhang2024memwarp} & 60.24 & 19.84 & 0.53 \\
textSCF \cite{chen2024textscf} & 60.75 & 22.44 & 0.87 \\
\hline
\textbf{SegReg-Dice(Ours)} & \textbf{68.02} & 20.77 & 0.58 \\
\textbf{SegReg-EDT (Ours)} & 67.19 & 21.04 & 0.52 \\
\textbf{SegReg-EDT\&Dice (Ours)} & 67.74 & 21.73 & 0.60 \\
%\hline
%\textbf{SegReg-Dice (Ours)} & \bf{93.24} & 10.23 & 0.56 \\
%\textbf{SegReg-EDT (Ours)} & 92.09 & \bf{9.62} & 0.48 \\
%\textbf{SegReg-EDT\&Dice (Ours)} & 92.73 & 10.08 & 0.55 \\
%Avg val dice 0.9209, Avg init dice 0.3086, Avg jac det 0.0030, Avg std dev 0.4821, Avg hd95 9.6213, Avg init hd95 29.7679
\hline
\end{tabular}
}
\caption{
Quantitative comparison on abdomen CT dataset. 
}
\label{tab:abdomen}
\end{center}
\vspace{-2ex}
\end{table}

\begin{table}[!ht]
\begin{center}
\resizebox{1.\linewidth}{!}{
\begin{tabular}{ lccc}
\hline
\rowcolor{lightgray}
Model & Dice (\%) $\uparrow$ & HD95 (mm) $\downarrow$ & SDlogJ $\downarrow$  \\ 
\hline
Initial & 58.14 & 11.95 & - \\
\hline
VoxelMorph \cite{balakrishnan2018unsupervised} & 81.95& 7.18& 0.07\\
VoxelMorph-Img\&Seg (GT) \cite{balakrishnan2018unsupervised} & 90.71 & 4.90 & 0.12 \\
VoxelMorph-Seg (GT)\cite{balakrishnan2018unsupervised} &91.32 &3.74 &0.08 \\
\hline
SegReg-Dice(GT)& 98.23& 1.93& 0.36 \\
SegReg-EDT(GT)& 95.44& 3.62& 0.08 \\
SegReg-EDT\&Dice(GT)& 98.19& 2.05& 0.14 \\
\hline
SegReg-Dice & 89.98 & 5.08 & 0.36 \\
SegReg-EDT & 87.79& 6.03 & 0.09 \\
SegReg-EDT\&Dice& 89.24& 5.45 & 0.15 \\
\hline
\end{tabular}
}
\caption{
Quantitative comparison between automatic and GT segmentation on the cardiac ACDC dataset. 
}\label{tab:GD-AUTO}
\end{center}
\vspace{-2ex}
\end{table}

\subsubsection{Inter-subject Registration on Abdomen Images.}
We also demonstrate the superior capability of our SegReg framework in managing much larger deformations through inter-subject abdominal CT registration, as quantitatively validated in Table \ref{tab:abdomen}. 
Consistent with findings on the ACDC dataset, both MemWarp \cite{zhang2024memwarp} and RDP \cite{wang_RDP} exhibit enhanced performance over conventional deep learning-based methods, attributable to their specialized design for large deformation scenarios. Notably, textSCF \cite{chen2024textscf}, which similarly incorporates segmentation masks in its registration network, achieves competitive registration accuracy. 
Provided with equivalent segmentation inputs as textSCF \cite{chen2024textscf}, our SegReg framework establishes new SOTA registration performance, attaining a mean Dice score of 68.02\%. This represents a substantial improvement of approximately 12\% over textSCF \cite{chen2024textscf}, while simultaneously improving the smoothness of deformation fields. %Furthermore, while SegReg-EDT shows a marginal decrease in Dice score (0.68\% reduction) compared to SegReg-Dice, it generates notably smoother deformation patterns. The hybrid SegReg-EDT&Dice variant achieves a balanced performance profile between the two base configurations, maintaining registration accuracy while improving field regularity.

\subsubsection{Comparison Between Automatic and GT Segmentation.}
%The aforementioned results are all predicted using the ground-truth segmentation. However, in realistic scenarios, ground-truth segmentation is not always available, and we have to use automatic segmentation instead. Therefore, we utilise two publicly available segmentation models, Swin-UNETR~\cite{tang2022self,hatamizadeh2022swin}(https://huggingface.co/darragh/swinunetr-btcv-base) and nnUNet~\cite{isensee2021nnu}(https://zenodo.org/records/3734294), to segment the abdominal CT image and cardiac MR images respectively, and fed their predictions into our SegReg framework. For both of them, we directly utilize their trained model online, and without any local fine-tuning. The resultant segmentation accuracy on our test abdominal images and ACDC test set are 75.83\% and 91.19\%, respectively. 

The presented SegReg performance is achieved using automated segmentation inputs with test Dice scores of 75.83\% (abdominal CT) and 91.19\% (ACDC). To further explore the theoretical upper bound, we implement SegReg with GT segmentation on ACDC (Table \ref{tab:GD-AUTO}). We develop two VoxelMorph variants: VoxelMorph-Seg (segmentation-only input) and VoxelMorph-Img\&Seg (image-segmentation concatenation). 
Comparative analysis reveals that integrating GT segmentation significantly enhances registration performance (about 10\% Dice improvement), with segmentation-only inputs outperforming combined image-segmentation inputs in segmentation-based evaluation. Using automated segmentations, SegReg's inherent capacity to preserve deformation field discontinuities enables comparable performance (p$>$0.05) to both VoxelMorph variants (feeding GT segmentation). With GT segmentation inputs, SegReg surpasses both variants, achieving a mean Dice score exceeding 95\% (SegReg-Dice and SegReg-EDT\&Dice are over 98\%). %This demonstrates that improving segmentation accuracy on SegReg can reach 98\% alignment.
\subsubsection{Iterative Registration on ThoraxCBCT.}
%Our SegReg framework can also be applied to iterative registration like traditional approaches, without pre-training prior to the registration. We demonstrate our SegReg on a multi-model intra-patient registration task on ThoraxCBCT, where we need to register the FBCT image of a subject into his CBCT images at the beginning and end of the therapy. We utilize the online version (https://totalsegmentator.com/) of Totalsegmentor \cite{d2024totalsegmentator} to segment the input images into 13 classes, and align them in each region. %add details about how to merge the segmentation labels in the appendixs
%As shown in Table \ref{tab:ThoraxCBCT}, our SegReg archives significantly better Dice score than the rest approaches. According to the leaderboard, our SegReg archives rank 2 among all the uploaded results. Especially on the HD score, the results of SegReg is 13.94mm, with about 13mm lower than the rank-1 approach. 

%fan-beam CT (FBCT) image into two cone-beam CT (CBCT)

Our SegReg framework is not only confined to learning-based registration but can also be adapted for iterative registration akin to conventional methods, negating the need for pre-training before registration. We exemplify the application of SegReg in multi-modal intra-patient registration in ThoraxCBCT, where the goal is to align a subject's fan-beam CT image with corresponding cone-beam CT images captured at the beginning and end of the therapy.
Leveraging the online Totalsegmentor \cite{d2024totalsegmentator}, we segment the input images into 13 distinct classes and proceed to align them regionally. As detailed in Table \ref{tab:ThoraxCBCT}, SegReg achieves a markedly superior Dice score compared to other approaches. Notably, the HD score of SegReg is 13.94mm, approximately 15mm lower than suboptimal approach.

\begin{table}[!t]
\begin{center}
\resizebox{\columnwidth}{!}{
\begin{tabular}{ lccccc } 
\hline
\rowcolor{lightgray} 
Model&  Dice (\%) $\uparrow$&  HD95(mm)& SDlogJ $\downarrow$\\ 
\hline
Initial & 31.3  & 55.358 & - \\ 
\hline 
deeds \cite{deeds2013Mattias} &  64.8&  29.03 & 0.15 \\
%LKFT-d1&62.9±4.6&12.134±2.790&48.500±5.126&0.174±0.014\\
%swinseg &59.5±4.6&15.555±2.958&50.640±5.274&0.050±0.015\\
ConvexAdam \cite{siebert2021fast} & 56.8 & 45.22 & 0.16 \\
ShiftMorph \cite{yang2024shiftmorph} & 58.4 & 53.19 & 1.26 \\
%deform\_whitaffine\_r3&57.7±4.4&14.791±3.325&52.634±3.378&1.205±0.487\\
%zeroshot&57.0±5.0&13.947±2.592&49.369±4.244&0.149±0.007\\
NiftyReg \cite{modat2010fast}& 56.8 & 50.08 & 0.06 \\
Fourier-Net \cite{jia2023fourier} & 54.8 & 52.32 & 0.09\\
%TR&53.1±4.8&15.572±3.307&52.877±3.046&1.181±0.497\\
VoxelMorph++ \cite{heinrich2022voxelmorph++}& 50.3 & 28.56 & 0.13\\
%fastreg\_v0&38.4±10.2&9.861±1.397&49.807±3.580&0.056±0.008\\
\hline
\textbf{SegReg (Ours)} &  \textbf{68.9} & \textbf{13.94} & 0.30 \\
\hline
\end{tabular}
}
\caption{
Quantitative comparison on the ThorexCBCT dataset, derived directly from the online leaderboard.
%Best-performing metrics are in bold. ``Initial" refers to baseline results before registration.
}\label{tab:ThoraxCBCT}
\end{center}
\vspace{-2ex}
\end{table}

\subsubsection{Discontinuity-preserving Analysis.}
%The final deformation fields of SegReg are composed of several sub-deformation fields, and therefore, the obtained deformation fields of SegReg is naturally discontinuity-preserving on the boundaries between different sub-regions. In addition, for each registration pair, the motion within each sub-region is smooth and continuous, guaranteed by smooth regularisation. Through this strategy, our SegReg can predict deformation fields which are smooth within each sub-region while discontinuity across the different regions, mimicking the realistic motion in 3D space. A visualisation of the motion vector can be found in Figure~\ref{fig:dice_vs_edt}. The discontinuity can be observed from both SegReg-Dice and SegReg-EDT, while not shown in the results of VoxelMorph. Comparing the results of SegReg-Dice and SegReg-EDT, we find that, the resultant deformation field of SegReg-EDT is smoother than SegReg-Dice, despite its registration accuracy being marginally lower. %In cardiac image registration, SgReg-DET can achieve over 95\% registration Dice and keep discontinuity on the boundary while keeping a similar global smooth registration than the rest global smooth registration approaches. SegReg-Dice, instead, has with a larger discontinuity (evidenced by the warped grid) on the boundary while higher registration accuracy.

\begin{figure}[ht]
\centering
\includegraphics[width=0.9\linewidth]{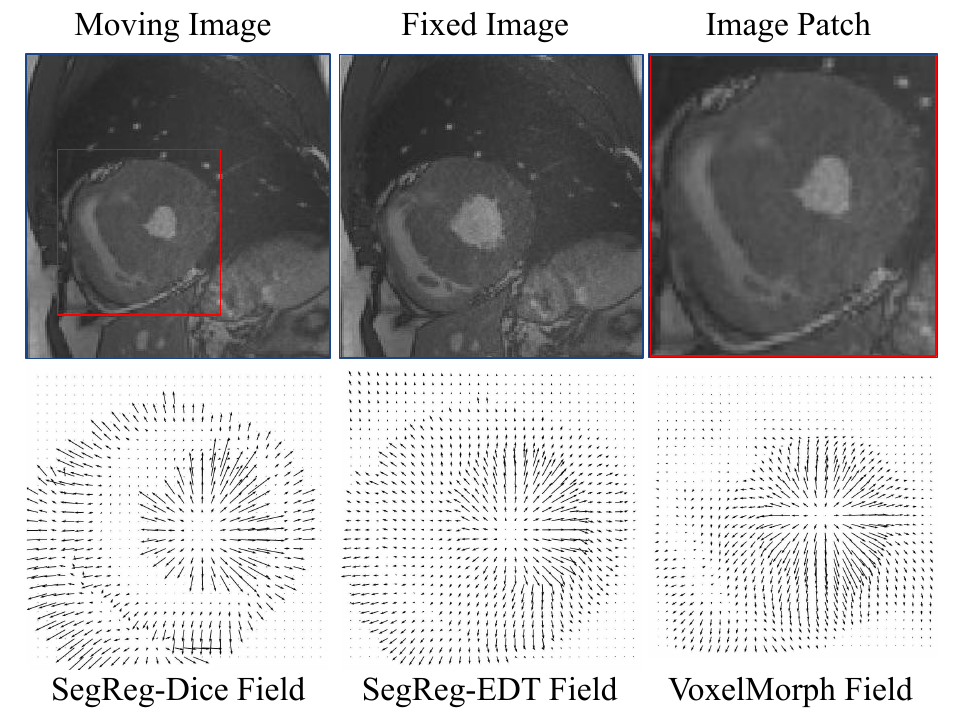}
\caption{Visualisation of the motion vector from deformation of SegReg-Dice, SegReg-EDT and VoxelMorph.% (Assumption A1). 
}
\label{fig:dice_vs_edt}
\vspace{-2ex}
\end{figure}

The final deformation fields of SegReg are generated as a composite of multiple sub-deformation fields, inherently preserving discontinuities at the boundaries between disparate sub-regions. For each registration pair, the motion within each sub-region maintains smoothness, ensured by smooth regularization. This approach enables our SegReg to predict deformation fields that are smooth within each sub-region and discontinuous across different regions, effectively emulating realistic motion. A visualization of the motion vectors is presented in Figure~\ref{fig:dice_vs_edt}. Discontinuities are evident in both SegReg-Dice and SegReg-EDT, in contrast to the results of VoxelMorph. Comparing SegReg-Dice and SegReg-EDT, the deformation field resulting from SegReg-EDT exhibits improved smoothness than that of SegReg-Dice.

\subsubsection{How Segmentation Affects Registration?}
%As SegReg mainly depends on the segmentation to align the moving and fixed images, the segmentation performance would limit the upper bound of the registration accuracy. To explore how the segmentation performance would affect the registration accuracy, we train a nnUNet from scratch and utilize the trained model at different training epochs to predict segmentation masks at different accuracies. Then we feed segmentations with different accuracy as the input of SegReg, and compare the resultant registration accuracy, with the corresponding results in Figure \ref{fig:seg_reg}. %Table.~\ref{tab: different_seg}.

\begin{figure}[ht]
\centering
\includegraphics[width=0.9\linewidth]{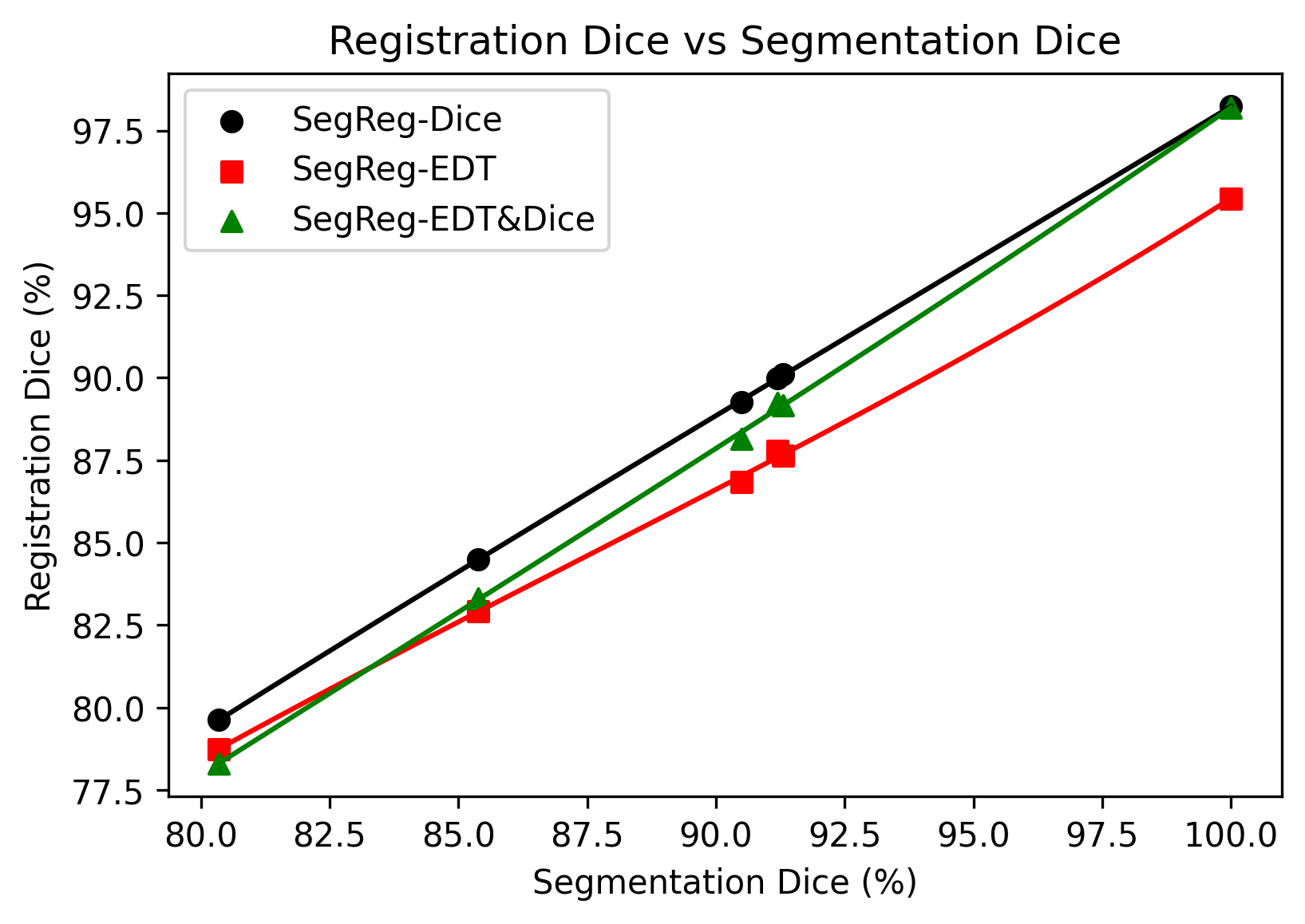}
\caption{The registration performance of SegReg with the increasing segmentation accuracy. 
}
\label{fig:seg_reg}
\vspace{-2ex}
\end{figure}

Given that SegReg primarily relies on segmentation to align the moving and fixed images, the quality of segmentation performance inherently sets an upper limit on the achievable registration accuracy. To investigate the impact of segmentation performance on registration accuracy, we train a nnUNet from scratch and employ the models at various training stages to generate segmentation masks with varying degrees of accuracy. Subsequently, we feed these different segmentations into SegReg and assess the resulting registration accuracy, as depicted in Figure \ref{fig:seg_reg}. 
We observed that \textbf{SegReg's registration accuracy shows a near-linear dependence on segmentation accuracy}. Although the registration accuracy of SegReg-EDT is slightly inferior to that of SegReg-Dice, it compensates by predicting smoother deformation fields. SegReg-EDT\&Dice gradually achieves a balance between these two models with increasing segmentation accuracy. The registration accuracy only experiences a marginal decline relative to the segmentation accuracy, demonstrating the validity of our segmentation-driven registration framework in preserving anatomical consistency. This finding reframes image registration as a segmentation-guided learning problem, where optimal correspondence is predicted automatically from precise tissue boundary delineation. %This finding repositions registration as a segmentation-guided manifold learning process, where the optimal correspondence emerges automatically when tissue boundaries are precisely delineated.

%Consequently, the registration task can be effectively transformed into a segmentation task, where optimal registration results can be obtained by pursuing accurate segmentation.
%This analysis elucidates the relationship between segmentation precision and the ensuing registration outcomes.

%It can be found that, the registration accuracy of SegReg can tightly follow the increase in segmentation accuracy, which is almost a linear correlation. The registration accuracy of SegReg-EDT is worse marginal worse than SegReg-Dice, while it can predict smoother deformation fields. The resultant registration accuracy is only with a marginal decrease to the segmentation accuracy, demonstrating that our idea to achieve registration by segmentation is reasonable. In this case, the registration task can be swift into a registration task, where an ideal segmentation mask can lead to ideal registration accuracy.

\begin{figure}[ht]
\centering
\includegraphics[width=0.9\linewidth]{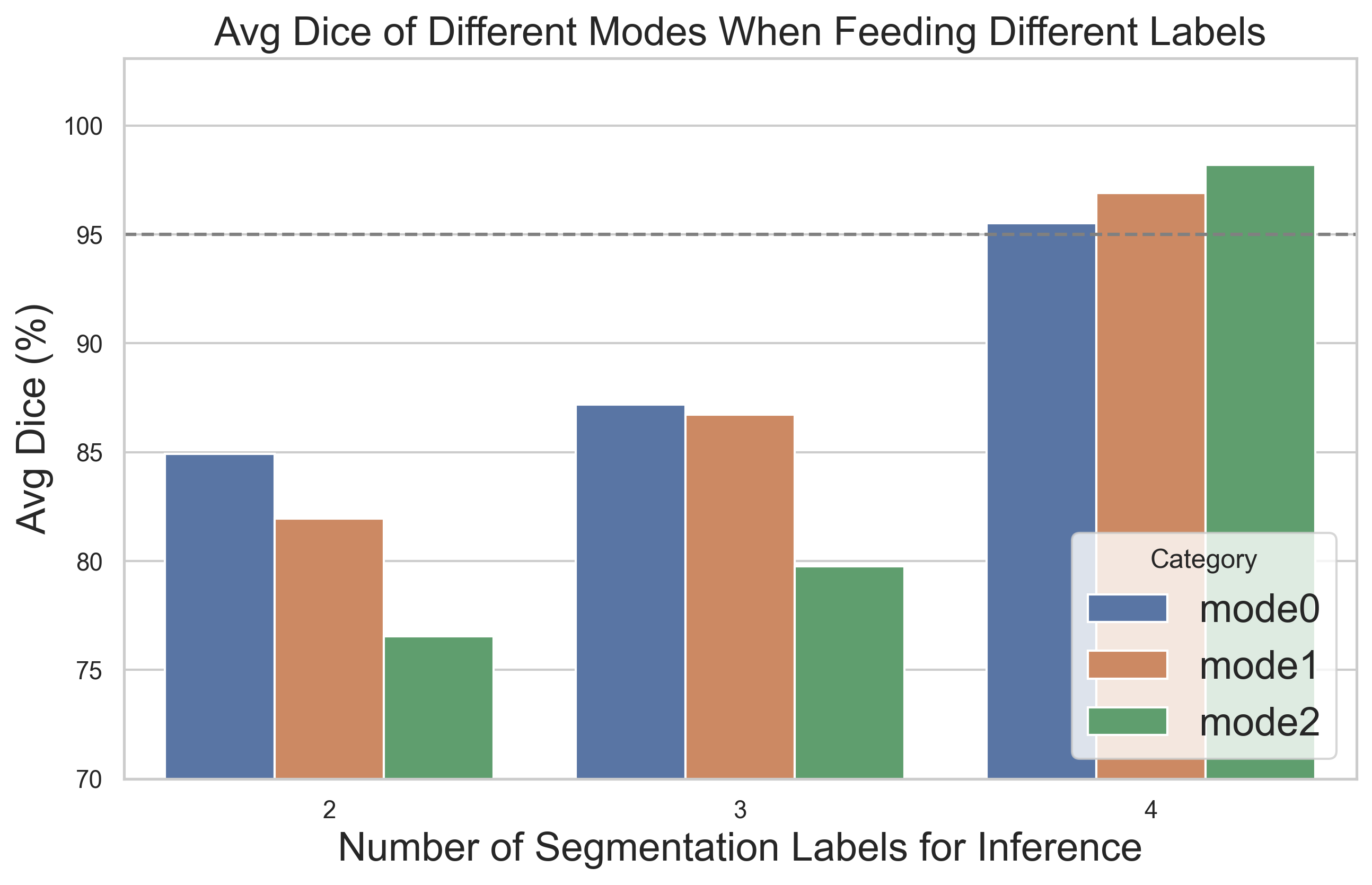}
\caption{The registration accuracy of different modes when fed different numbers of labels as inputs. 
}
\label{fig:mode_dice}
\vspace{-3ex}
\end{figure}

\subsubsection{Influence of the Number of Segmentation Classes.} %retrain the network?
%The segmentation classes determine how many regions the original images would be divided into. The selection of the number of segmentation classes is important, as too few segmentation labels would lead to low registration accuracy, and too many segmentation labels would significantly increase the complexity of the registration. To investigate its influence on registration accuracy, we gradually merge some segmentation labels and fed updated segmentation for inference (using the same trained model as Table \ref{tab:ACDC}), with the corresponding results in Table.~\ref{tab: different_num_seg}. 

%Insufficient segmentation labels may result in diminished registration accuracy, while an excess could substantially amplify the complexity of registration.

%The number of segmentation classes dictates the division of the original images into distinct regions, a decision critical to the success of the registration process. 
To elucidate the impact of number of segmentation classes on registration precision, we systematically merge some segmentation labels and feed the revised segmentations for training, leading to three modes: mode0 (foreground and background), mode1 (LV (LVBP+LVM), RV and background) and mode2 (LVBP, LVM, RV and background). 
In inference, we also introduce three distinct label configurations to evaluate the trained models, segmenting the original images into two (foreground+background), three (background+LV+RV), and four regions (background+LVBP+LVM+RV), as shown in Figure \ref{fig:mode_dice} (details in Appendix). The registration accuracy increases with a higher number of specified regions during inference, irrespective of the training labels. Interestingly, when comparing the outcomes of mode0, mode1, and mode2 using four regions (LVBP, LVM, RV, and background) as inputs, we observe no significant discrepancies, all exceeding 95\% Dice. Hence, we conclude that within the SegReg framework, registration precision is predominantly determined by the number of regions defined for inference, rather than a strict correspondence between training and testing labels.

\section{Conclusion}
This work introduces a segmentation-enhanced registration framework, SegReg, which substantially surpasses current performance limits in both mono- and multi-modal registration. Unlike global smoothness constraints, SegReg predicts anatomically guided regional deformation fields via segmentation, preserving tissue boundary discontinuities. SegReg achieves near-perfect alignment (98.23\% Dice) with GT segmentations while maintaining state-of-the-art performance even with automated segmentation inputs. Crucially, SegReg reveals a near-linear dependence of registration accuracy on segmentation quality, effectively transforming registration into a segmentation-driven process. Future iterations could replace the basic UNet backbone with more efficient backbones, and integrate foundation models like SAM to address segmentation challenges.

\bibliography{aaai2026}

\appendix
\renewcommand{\thefigure}{A\arabic{figure}} 
\renewcommand{\thetable}{A\arabic{table}} 
\setcounter{figure}{0} 
\setcounter{table}{0}

\clearpage
\section{Appendix}
\subsubsection{A: Details of Registration in Abdomen Image.}

% \subsubsection{Abdomen CT Dataset}
% To demonstrate the effectiveness of our EOIR framework, we conducted inter-subject registration on the Abdomen CT dataset \cite{xu2016evaluation} from the Learn2Reg challenge \cite{hering2022learn2reg} 2020. 
Compared with intra-subject registration, inter-subject registration is generally more challenging, due to the great variability and large deformations across subjects. 
We utilize a dataset of 30 abdominal CT scans \cite{xu2016evaluation} from the Learn2Reg challenge, each accompanied by segmentation masks for 13 anatomical structures. 
The dataset is divided into training, validation, and test sets, with 20, 3, and 7 scans, respectively, resulting in 380 ($20\times19$) training pairs, 6 ($3\times2$) validation pairs, and 42 ($7\times6$) testing pairs. 
All images were resampled to a voxel resolution of 2 mm and standardized to a spatial size of $192\times160\times256$.

\begin{figure*}[htb]
\centering
\includegraphics[width=\linewidth]{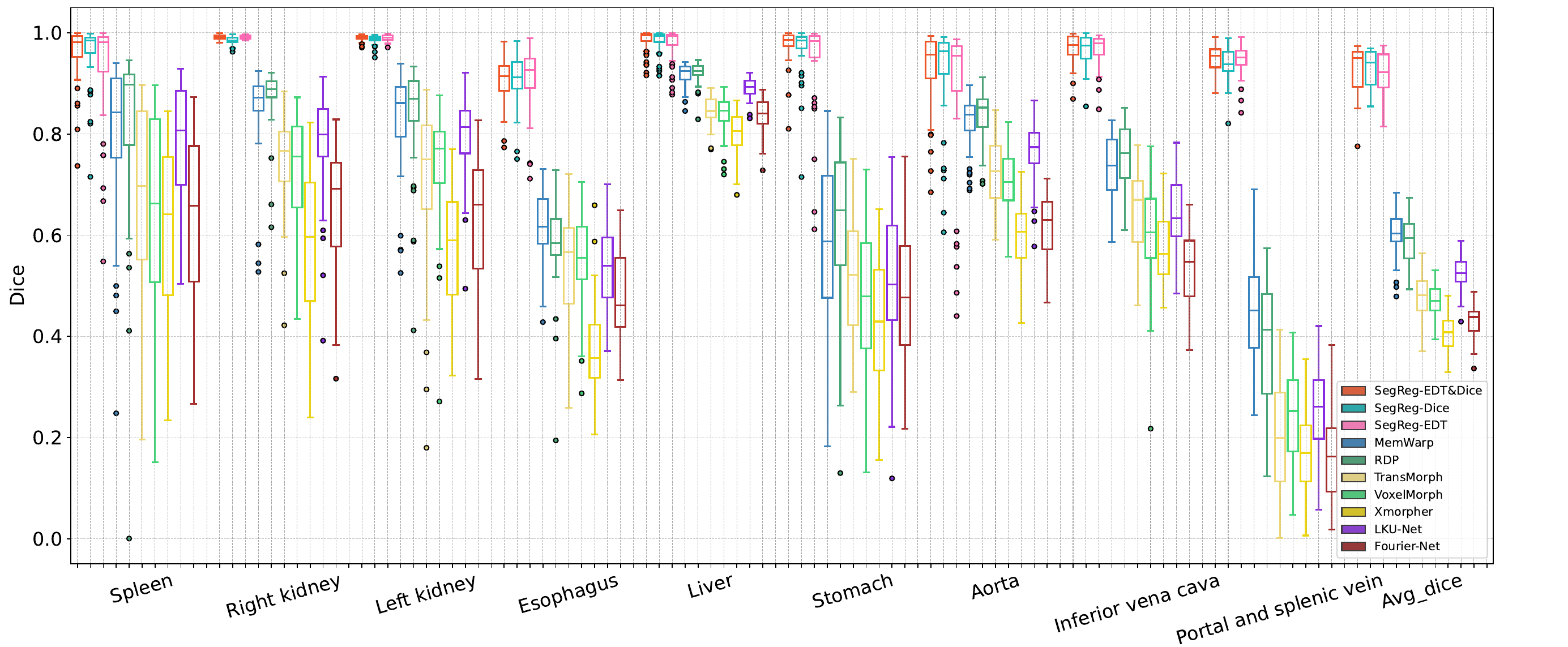}
\caption{Boxplots on the abdomen CT dataset. SegReg-Dice, SegReg-EDT and SegReg-EDT\&Dice achieve significantly higher Dice than the rest of the methods.
}
\label{fig:boxplot_abdomen}
\end{figure*}

We utilzie an online model {Swin-UNETR}~\cite{tang2022self,hatamizadeh2022swin} to segment the abdomen images for our SegReg, available at \url{https://huggingface.co/darragh/swinunetr-btcv-base}.
The registration results of SegReg when using segmentation as inputs are presented in Table \ref{tab:abdomen_GT}. The distribution of results on abdomen registration is plotted in Figure \ref{fig:boxplot_abdomen}, corresponding to the results in Table \ref{tab:abdomen} and Table \ref{tab:abdomen_GT}.

\begin{table}[!htb]
\begin{center}
\resizebox{\columnwidth}{!}{
\begin{tabular}{ lccc }
\hline
\rowcolor{lightgray}
Model &  Dice (\%) $\uparrow$ & HD95 (mm) $\downarrow$ & SDlogJ $\downarrow$ \\ 
\hline
Initial & 30.86 & 29.77 & - \\
% ConvexAdam~\cite{siebert2021fast} & 84.64 & 1.50 & \textbf{0.07} \\
% \hline
% VoxelMorph \cite{balakrishnan2018unsupervised} & 47.05 & 23.08 & 0.13\\
% TransMorph \cite{chen2022transmorph} & 47.94 & 21.53 & 0.13 \\
% LKU-Net \cite{jia2022u} & 52.78 & 20.56 & 0.98 \\
% Xmorpher \cite{shi2022xmorpher} & 40.67& 23.77&0.14\\
% LapIRN \cite{mok2020large} & 54.55 & 20.52 & 1.73 \\
% %LessNet \cite{jia2024decoder} & 42.03 & 27.03 & \bf{0.07} \\
% %FourierNet \cite{jia2023fourier} & 48.49 & 20.69 & 0.13 \\
% Fourier-Net \cite{jia2023fourier} & 42.80 & 22.95 & 0.13 \\
% ConvexAdam \cite{siebert2021fast} & 51.10 & 23.14 & \textbf{0.11}\\ 
% SAMConvex \cite{li2023samconvex} & 53.65 & \textbf{18.66} & 0.12 \\
% \hline
% RDP \cite{wang_RDP} & 58.77 & 20.07 & 0.22 \\
% MemWarp \cite{zhang2024memwarp} & 60.24 & 19.84 & 0.53 \\
% textSCF \cite{chen2024textscf} & 60.75 & 22.44 & 0.87 \\
% \hline
\textbf{SegReg-Dice(Ours)} & 68.02 & 20.77 & 0.58 \\
\textbf{SegReg-EDT (Ours)} & 67.19 & 21.04 & 0.52 \\
\textbf{SegReg-EDT\&Dice (Ours)} & 67.74 & 21.73 & 0.60 \\
\hline
\textbf{SegReg-Dice (GT)} & \bf{93.24} & 10.23 & 0.56 \\
\textbf{SegReg-EDT (GT)} & 92.09 & \bf{9.62} & 0.48 \\
\textbf{SegReg-EDT\&Dice (GT)} & 92.73 & 10.08 & 0.55 \\
% Avg val dice 0.9209, Avg init dice 0.3086, Avg jac det 0.0030, Avg std dev 0.4821, Avg hd95 9.6213, Avg init hd95 29.7679
\hline
\end{tabular}
}
\caption{
Quantitative comparison between the registration performance of automatic segmentation and GT segmentation on the abdomen CT dataset. 
}
\label{tab:abdomen_GT}
\end{center}
\end{table}

\subsubsection{B: Details of Registration on ACDC.}
% \subsubsection{ACDC Dataset} %MSE 1:0.01
We evaluate our SegReg in intra-subject cardiac MR image registration using the ACDC dataset \cite{bernard2018deep}, with 80 subjects for training, 20 for validation, and 50 for testing. 
For each subject, the end-diastole (ED) and end-systole (ES) images have GT segmentations for LVBP, LVM and RV. 
We perform registration from both ED to ES and ES to ED, resulting in 160 training pairs, 40 validation pairs, and 100 test pairs. 
All images are pre-processed through resampling, normalization, cropping, and padding, into $128 \times 128 \times 16$ (spacing:$1.8 \times 1.8 \times 10 mm^3$).

\begin{figure*}[ht]
    \centering
    \includegraphics[width=1.0\linewidth]{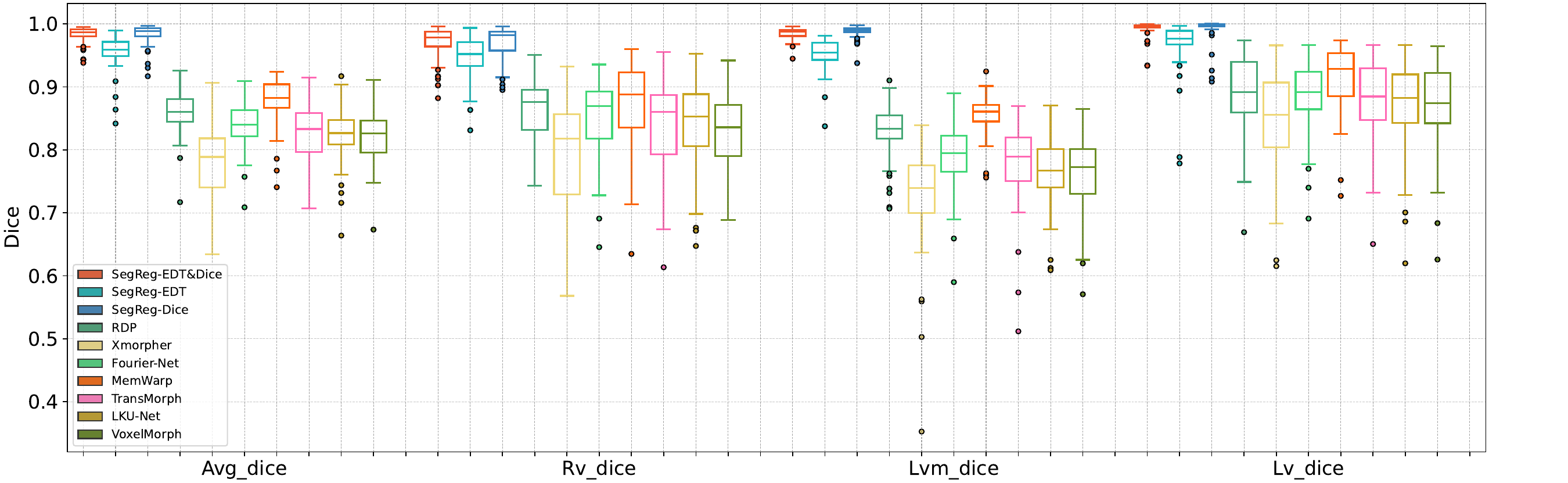}
    \caption{
    Boxplots on the ACDC dataset. SegReg-Dice, SegReg-EDT and SegReg-EDT\&Dice achieve significantly higher Dice than the rest of the methods.
    }
    \label{fig:acdc_boxplot}
\end{figure*}

We utilize an online model {nnUNet}~\cite{isensee2021nnu} to segment the cardiac images for our SegReg, available at \url{https://zenodo.org/records/3734294}.
The distribution of registration results on ACDC is presented in Figure \ref{fig:acdc_boxplot}, corresponding to the results in Table \ref{tab:ACDC}. In addition, the detailed results of Figure \ref{fig:seg_reg} are shown in Table \ref{tab: different_seg}. To obtain labels with varying segmentation, we train a nnUNetV2 from scratch, and then feed the results of different epochs (different accuracy) into SegReg, as noted in the table. The `nnUnetV1-online' is the trained model we obtain directly online, without any fine-tuning, whose segmentation performance is only marginally lower than the best model trained by ourselves (``nnUnetV2-Best(997)"). Table \ref{tab:different_num_seg} presents the detailed results in Figure \ref{fig:mode_dice}.

\begin{table*}[ht]
\begin{center}
\resizebox{1.\linewidth}{!}{
\begin{tabular}{ lccccccc }
\hline
\hline
\rowcolor{lightgray}
Model & Seg Dice (\%) $\uparrow$ & Avg. Dice (\%) $\uparrow$ & LV Dice (\%) $\uparrow$ & LVM Dice (\%) $\uparrow$ & RV Dice (\%) $\uparrow$ & HD95 (mm) $\downarrow$ & SDlogJ $\downarrow$  \\ 
\hline
Initial & 58.14 & 61.60 & 48.33 & 64.50 & 11.95 & - &-\\
\hline
SegReg-Dice (GT Seg) &100.00 &98.23 &99.15 &98.74 &96.81 &1.93 &0.36 \\
SegReg-Dice (nnUnetV2-1) &80.35 &79.62 &84.77 &76.15 &77.94 &8.84 &0.37 \\
SegReg-Dice (nnUnetV2-2) &85.38 &84.49 &89.09 &80.88 &83.52 &6.91 &0.31 \\
SegReg-Dice (nnUnetV2-11) &90.50 &89.27 &92.92 &87.25 &87.62 &5.19 &0.34 \\
SegReg-Dice (nnUnetV2-Best(997)) &91.31 &90.11 &93.05 &88.20 &89.08 &4.83 &0.35 \\
SegReg-Dice (nnUnetV1-online) &91.19 &89.98 &92.85 &88.18 &88.91 &5.08 &0.36 \\
\hline
SegReg-EDT (GT Seg) &100.00 &95.44 &96.69 &95.17 &94.46 &3.62 &0.07 \\
SegReg-EDT (nnUnetV2-1) &80.35 &78.74 &83.49 &76.70 &76.03 &8.61 &0.14 \\
SegReg-EDT (nnUnetV2-2) &85.38 &82.93 &87.46 &79.61 &81.70 &7.53 &0.12 \\
SegReg-EDT (nnUnetV2-11) &90.50 &86.83 &90.79 &84.81 &84.89  &6.24 &0.10 \\
SegReg-EDT (nnUnetV2-Best(997)) &91.31 &87.64 &90.92 &85.32 &86.69 &5.95 &0.08 \\
SegReg-EDT (nnUnetV1-online) &91.19 &87.79 &90.89 &85.67 &86.81 &6.03 &0.09 \\
\hline
\hline
\end{tabular}
}
\end{center}
\caption{
Registration performance of SegReg with the increasing segmentation on the cardiac ACDC dataset, where nnUnetV2-1, nnUnetV2-2, nnUnetV2-11 and nnUnetV2-Best(997) mean the segmentation are predicted using the traiend nnUNet model at 1, 2, 11 and 997 epochs, respectively.
Symbols indicate direction: $\uparrow$ for higher is better, $\downarrow$ for lower is better. 
``Initial" refers to baseline results before registration.
}\label{tab: different_seg}
\end{table*}

\begin{table*}[ht]
\begin{center}
\resizebox{1.\linewidth}{!}{
\begin{tabular}{ lcccccc }
\hline
\hline
\rowcolor{lightgray}
Model & Avg. Dice (\%) $\uparrow$ & LV Dice (\%) $\uparrow$ & LVM Dice (\%) $\uparrow$ & RV Dice (\%) $\uparrow$ & HD95 (mm) $\downarrow$ & SDlogJ $\downarrow$  \\ 
\hline
Initial & 58.14 & 61.60 & 48.33 & 64.50 & 11.95 & - \\
\hline
mode0 (background+foreground) & 84.91 & 82.65 & 78.87 & 93.19 & 5.81 & 0.07\\
mode0 (background+LV+RV) & 87.19 & 82.52 & 82.19 & 96.87 & 5.19 & 0.07 \\
mode0 (background+LVBP+LVM+RV) & 95.52 & 95.03 & 94.74 & 96.78 & 3.24 & 0.11 \\
\hline
mode1 (background+foreground) & 81.94 & 80.23 & 75.36 & 90.24 & 6.96 & 0.06\\
mode1 (background+LV+RV) & 86.72 & 81.55 & 81.90 & 96.70 & 5.54 & 0.08 \\
mode1 (background+LVBP+LVM+RV) & 96.89 & 97.35 & 96.65 & 96.67 & 3.14 & 0.10 \\
\hline
mode2 (background+foreground) & 76.54 & 71.41 & 66.58 & 91.62 & 8.06 & 0.06\\
mode2 (background+LV+RV) & 79.76 & 71.52 & 70.82 & 96.94 & 7.00 & 0.07\\
mode2 (background+LVBP+LVM+RV) & 98.18 &99.21 &98.44 & 96.90& 2.05 & 0.14\\
\hline
\hline
\end{tabular}
}
\end{center}
\caption{
Registration results of feeding different types of segmentation components into the SegReg framework.
Symbols indicate direction: $\uparrow$ for higher is better, $\downarrow$ for lower is better. 
``Initial" refers to baseline results before registration.
}\label{tab:different_num_seg}
\end{table*}

\subsubsection{C: Details of Registration in ThoraxCBCT.}
\label{sec:rationale}

% %todo
% \subsubsection{ThoraxCBCT} %MSE 1:0.01
To demonstrate the iterative registration performance of our SegReg framework, we conduct another intra-subject registration work using CT images from ThoraxCBCT \cite{hugo2017longitudinal} in the Learn2Reg2023 challenge, where the registration is to register a planning fan-beam CT (FBCT) image into two cone-beam CT (CBCT) images at the beginning and end of therapy. The lack of paired data and the difference in image quality and resolution make it a challenging task. ThoraxCBCT contains three subjects for evaluation on the leaderboard. All the images are preprocessed by the challenge organiser to $390 \times 280 \times 300$ with a spacing of $1 \times 1 \times 1 mm^3$.

To register the thorax images from ThoraxCBCT, all the testing data are segmented using \url{https://totalsegmentator.com/}{Totalsegmentator}. As there are a total of 117 segmentation labels and many of them are not correlated to the regions/organs on the thorax, we select and merge the segmentation labels into 13 classes, comprising aorta, inferior vena cava, lung, vertebrae and ribs, esophagus, trachea, pulmonary artery, left scapula, right scapula, left clavicula, right clavicula, heart and background.

\end{document}